\documentclass[11pt]{article}
\usepackage{coling2020}
\usepackage{times}
\usepackage{graphicx}
\usepackage{url}
\usepackage{latexsym}
\usepackage{amsmath}
\usepackage{float}
\usepackage{hyperref}
\usepackage{url}
\usepackage{xspace}
\usepackage{multirow}
\usepackage{booktabs}
\usepackage{caption}
\usepackage{subfigure}
\usepackage{array}
\usepackage{wrapfig}

\usepackage{amssymb}
\usepackage{comment}
\usepackage{color}

\newcommand{\ie}{\emph{i.e.,}\xspace}
\newcommand{\eg}{\emph{e.g.,}\xspace}

\usepackage{color}

\usepackage{fdsymbol}

\colingfinalcopy 

\title{Context Modeling with Evidence Filter for Multiple Choice Question Answering}

\author{Sicheng Yu$^\clubsuit$,~~Hao Zhang$^{\vardiamondsuit,\varheartsuit}$,~~Wei Jing$^{\spadesuit,\varheartsuit}$,~~Jing Jiang$^\clubsuit$ \\
$^\clubsuit$School of Information System, Singapore Management University, Singapore \\
$^\vardiamondsuit$School of Computer Science and Engineering, Nanyang Technological University, Singapore \\
$^\spadesuit$Institute of High Performance Computing, A*STAR, Singapore \\
$^\varheartsuit$Institute for Infocomm Research, A*STAR, Singapore\\
{\tt \{scyu.2018@phdcs.,jingjiang@\}smu.edu.sg,\{26hzhang,21wjing\}@gmail.com}
}

\date{}

\begin{document}
\maketitle
\begin{abstract}
Multiple-Choice Question Answering (MCQA) is a challenging task in machine reading comprehension. A main challenge in MCQA is to extract ``evidence'' from the given context that supports the correct answer. In OpenbookQA dataset~\cite{mihaylov2018can}, the requirement of extracting ``evidence'' is particularly important due to the mutual independence of sentences in the context. Existing work tackles this problem by annotated evidence or distant supervision with rules which overly rely on human efforts. To address the challenge, we propose a simple yet effective approach termed \textbf{evidence filtering} to model the relationships between the encoded contexts with respect to different options collectively, and to potentially highlight the evidence sentences and filter out unrelated sentences. In addition to the effective reduction of human efforts of our approach compared, through extensive experiments on OpenbookQA, we show that the proposed approach outperforms the models that use the same backbone and more training data; and our parameter analysis also demonstrates the interpretability of our approach. 
\end{abstract}

\section{Introduction}\label{introduction}
Multiple-Choice Question Answering (MCQA) is a natural language processing task that has been attracting much attention recently due to its wide range of applications. One of the key challenges for MCQA is that the evidence sentences, which support the correct answer, are overwhelmed by a large number of unrelated sentences in the context. This is more serious in OpenbookQA dataset~\cite{mihaylov2018can} as the context of OpenbookQA consists of several independent facts and distribution of evidence sentences is random, which is difficult to locate the evidence sentence precisely. To address this issue, one solution is to train the model to learn how to extract evidence from context~\cite{yang2018hotpotqa,Xiao2019DynamicallyFG,Ding2019CognitiveGF}. However, it requires a large amount of annotated data, which is not practical in real-world scenarios. Another stream of work~\cite{wang2019evidence} utilizes distant supervision with a series of human-designed rules, which still requires much human effort.

\begin{figure}[b!]
    \centering
  \includegraphics[width=0.98\textwidth]{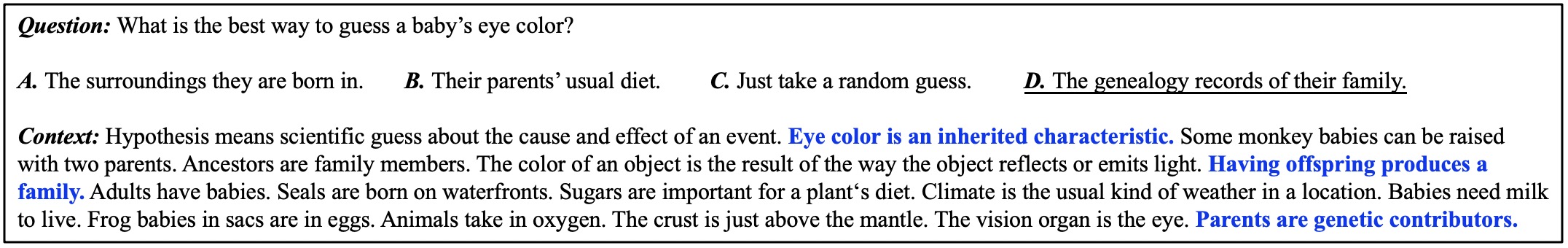}
  \caption{\small An example in OpenbookQA. Sentences in blue and bold are the evidence while the correct option is \underline{underlined}.}
  \label{example}
\end{figure}

In this paper, we present a simple yet effective approach, termed evidence filter, to potentially filter out irrelevant context and highlight the evidence without any human intervention. The key idea is to model the relationships between the encoded contexts with respect to different options. Specifically, we observe that existing methods for multiple-choice question answering typically encode each option with the context independently~\cite{sun2018improving,devlin2018bert}.  

Our method is based on the following observations and assumptions: (1) If a sentence in the context has a similar level of relevance on all of the given options, then it is highly likely that this sentence is not useful for answering the question. For instance, the sentence \textit{``The color of an object is the result of the way the object reflects or emits light.''} is not relevant to any of the options in Figure~\ref{example}. We therefore believe that this sentence is unlikely to be an evidence sentence. (2) An evidence sentence in the context is likely to be closely related to the correct option but irrelevant to the incorrect options. For instance, the evidence sentences shown in blue are indeed more related to the ground-truth option $\textit{D}$ than to the other options. Motivated by the aforementioned assumptions, we propose a method to capture the differences among context sentences with respect to the options via a carefully designed evidence filter, which can be treated as a denoising process or a high-pass filter. 

In a nutshell, we propose a simple yet effective method termed evidence filter to potentially extract evidence from context, and the parameters of evidence filter are interpretable. Experiment results on OpenbookQA dataset demonstrate the effectiveness of our approach, \eg, outperforms BERT Multi-task by $1.8\%$ with the same backbone and less training data.

\section{Methodology}\label{model}

\begin{wrapfigure}{r}{0.55\textwidth}
\vspace{-1cm}
    \centering
    \includegraphics[width=0.54\textwidth]{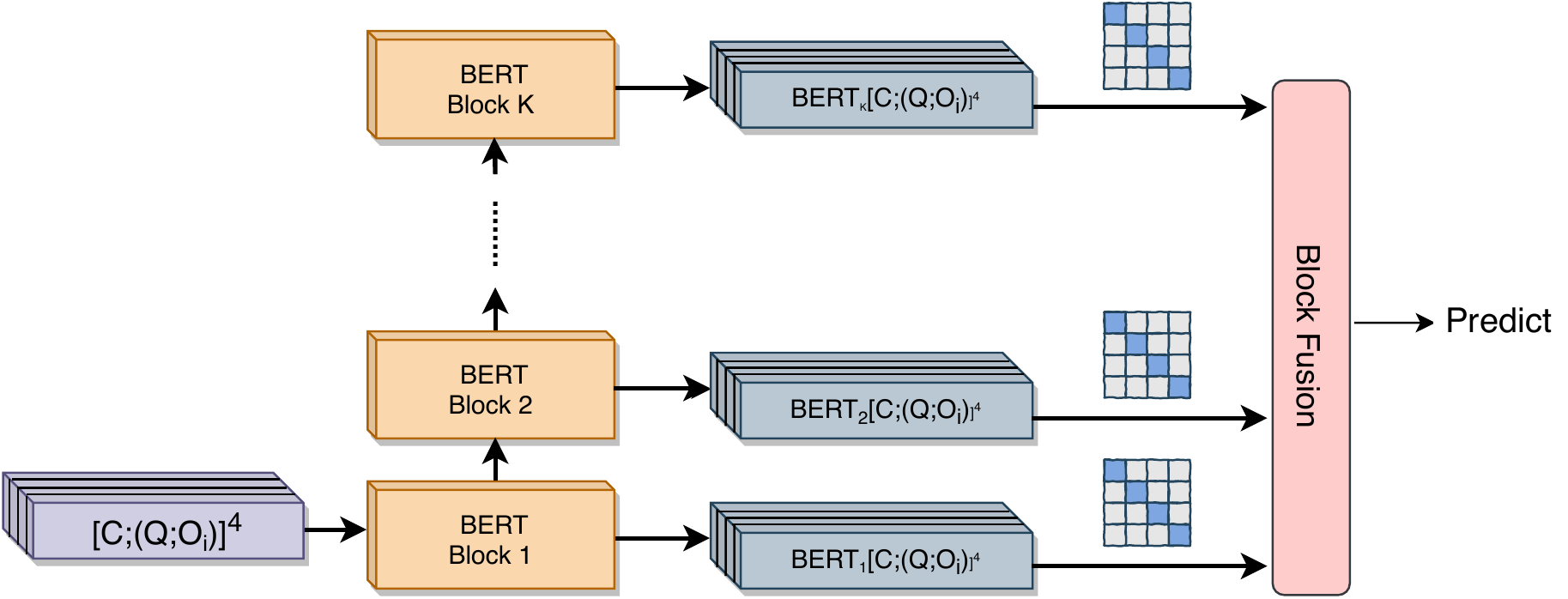}
	\caption{An overview of our proposed model.}
	\label{architecture}
	\vspace{-0.5cm}
\end{wrapfigure}

Our proposed framework consists of two components: 1) an \emph{evidence extraction} module that uses pre-trained BERT to process the context $C$, the question $Q$ and one of the options $O_i$ in order to implicitly extract evidence from $C$ that justifies $O_i$ as the answer to $Q$; 2) an \emph{evidence filter} module that adjusts the extracted evidence representation by considering the relationship between the evidence extracted from the previous module with respect to different options. Figure~\ref{architecture} depicts the overall architecture of the proposed framework.

\subsection{Evidence Extraction} 
The goal of the evidence extraction module is to process the context with respect to the question and a particular option in order to obtain a vector that represents the evidence from the context justifying the option. Suppose we have the context $C$ (which consists of a list of sentences extracted from a given corpus by an information retrieval method), question $Q$ and four answer options $(O_1,O_2,O_3,O_4)$. The evidence extraction module uses BERT~\cite{devlin2018bert} to process the sequence $[C; (Q; O_i)]$ for $i\!=\!1, 2, 3, 4$, where $C$ is treated as the first segment and the concatenation of $Q$ and $O_i$ is treated as the second segment for BERT to process. Since BERT has multiple blocks (\ie, layers), we use $\text{BERT}_k([C; (Q; O_i)]) \in \mathbb{R}^d$ to denote the output vector (after pooling) from the $k$-th block of BERT after processing context question and $i$-th option. With 4 options, we use $\text{BERT}_k([C; (Q; O_i)]^4) \in \mathbb{R}^{4 \times d}$ to denote all 4 output vectors.
We adopt BERT-large with $K\!=\!24$ blocks due to its better performance and generalization ability compared to BERT-base. 

\subsection{Evidence Filter}\label{sec:interaction_matrix}
Recall that in Section~\ref{introduction} we pointed out that whether or not a sentence serves as evidence to support an option also depends on whether this sentence is relevant to other options. Our assumptions are that (1) sentences related to the 4 options in similar ways are unlikely to be useful evidence, and (2) sentences related to one option but not other options are likely to be useful evidence. Based on these assumptions, we introduce an evidence filter matrix to adjust the extracted evidence representations from the \emph{evidence extraction} module; specifically, the evidence filter matrix will reduce the importance of evidence that is equally relevant to the 4 options, and subsequently place more emphasis on evidence relevant to a particular option.

Specifically, this is a $4\times4$ matrix inspired by~\cite{yu2018modelling}, denoted as $\mathbf{A}$, that will be applied to $\text{BERT}_k([C; (Q; O_i)]^4)$. Its diagonal values represent how much we want to maintain the originally extracted evidence representation for each option, whereas off-diagonal values represent how much reduction we would like to incur based on relations across different options. We thus expect the signs of the diagonal values to be the opposite of the signs of the off-diagonal values in $\mathbf{A}$. After the adjustment, the new evidence representations will be $\mathbf{A} \cdot \text{BERT}_k([C;(Q;O_{i})]^4) $.

In the early experiments, we observe that random initialization of each entry in the evidence filter individually may lead to inconsistent predictions when the 4 answer options are shuffled, i.e., when the order of the options changes. To alleviate such drawback, we apply the following constraints to the evidence filter. Specifically, we constrain the evidence filter matrix such that all its diagonal entries are the same, denoted as $\alpha$, and all its off-diagonal entries are also the same, denoted as $\beta$\footnote{Some justification and derivation of this constraint is provided in the supplementary material.}. They are represented as the blue cells and grey cells in the evidence filter matrix shown in Figure~\ref{architecture}.

With this constrained evidence filter matrix, the representation of evidence for the 1st option after this filter, for example, is $\alpha\cdot\text{BERT}_k([C;(Q;O_{1})]) + \sum_{i\neq1}\beta\cdot\text{BERT}_k([C;(Q;O_{i})])$, which exactly extracts the difference when the values of $\alpha$ and $\beta$ have opposite signs. It helps the model to highlight the evidence through taking the differences of the extracted contexts with respect to different options. 

We also adopt different evidence filters $\mathbf{A}_k$ for each block in BERT due to its better expression than sharing one evidence filter among all blocks.  Meanwhile, it also makes the block fusion layer more explicit.\footnote{The same pattern of evidence filter in Figure~\ref{architecture} is for simplicity.} A residual connection is adopted after the evidence filter followed by layer normalization~\cite{lei2016layer}. The intermediate output, $\text{BERT}_k'([C;(Q;O_{i})]^4)$, for block $k$ before block fusion is computed by:
\begin{equation}
    \text{BERT}_k'([C;(Q;O_{i})]^4)=\text{LayerNorm} \left ( \text{BERT}_k([C;(Q;O_{i})]^4)
    +\mathbf{A}_k \cdot \text{BERT}_k([C;(Q;O_{i})]^4) \right).
\end{equation}

Then a block fusion layer is adopted to integrate the intermediate output from each block by a single linear layer, and the output of block fusion layer $\mathbf{M}$ is defined as: 
\begin{equation}
    \mathbf{M}=\mathbf{W}_{\text{bf}}\Big(\big[\text{BERT}_1'([C;(Q;O_{i})]^4); \ldots; \text{BERT}_K'([C; (Q;O_{i})]^4)\big]\Big)+\mathbf{b}_{\text{bf}},
    \label{block fusion}
\end{equation}
where output is $\mathbf{W}_{\text{bf}}\in{\mathbb{R}^{1\times{K}}}$, $\mathbf{b}_{\text{bf}}\in{\mathbb{R}^{1}}$ and $\mathbf{M}\in{\mathbb{R}^{4\times{d}}}$. 

Finally, the model makes prediction after a linear layer and the standard cross-entropy loss is utilized as the loss function. Note that, compared to the BERT model, our method only requires a few more parameters, \ie, those for the evidence filters and those at the fusion layer.

\section{Experiments}\label{experiment}

\subsection{Datasets and Implementation}

\paragraph{Datasets.} To evaluate our model, we conduct experiments on a challenging MCQA dataset, OpenbookQA~\cite{mihaylov2018can}, which contains around 6000 $4$-way multiple-choice science questions and a corpus including roughly 7300 facts. Compared to other MCQA datasets, OpenbookQA is more complicated since the context in this dataset may not contain the answers explicitly~\cite{chen2019understanding}.

\paragraph{Experimental Settings.}
Following~\cite{sun2018improving}, we utilize PyLucene~\cite{bialecki2012apache} to retrieve the top-30 ranked sentences from the corpus containing the facts\footnote{We mainly focus on the method after the retrieval step. Thus, the information retrieval module is not included in this section.
One can easily adapt our model with any information retrieval methods.}, 

then we fine-tune BERT-large on RACE~\cite{lai2017race} with the na\"ive model as in~\cite{devlin2018bert}. Finally we train the model on OpenbookQA following the architecture in Figure~\ref{architecture}. The batch size is set to 32 and learning rate is $1e-5$, and linear warm up strategy is used with the first 10\% of the whole training steps. We also add another group of input, $[Q;O_{i}]$ sharing the parameters of BERT but evidence filter are not used for this group of input. The reason is that BERT contains commonsense knowledge which is helpful in scientific questions in OpenbookQA dataset. 

\begin{table}[hbtp]
\centering
\small
\begin{tabular}{l c c}
\toprule
\multirow{2}{*}{Methods}             & \multicolumn{2}{c}{OpenbookQA} \\ 
                                                      & Dev (\%)       & Test (\%)      \\ 
\midrule
Question Match + ELMo~\cite{mihaylov2018can}          & 54.6           & 50.2           \\ 
Odd-one-out Solver~\cite{mihaylov2018can}             & 56.9           & 50.2           \\ 
ESIM + ELMo~\cite{mihaylov2018can}                    & 53.9           & 48.9           \\ 
OFT~\cite{sun2018improving}                           & -              & 52.0           \\ 
OFT (ensemble)~\cite{sun2018improving}                & -              & 52.8           \\ 
Reading Strategies+GPT~\cite{sun2018improving}            & -              & 55.2           \\ 
Reading Strategies+GPT (ensemble)~\cite{sun2018improving} & -              & 55.8           \\ 
BERT-large (leaderboard)                                           & -              & 60.4           \\ 
BERT(large) Multi-task (leaderboard)                                      & -              & 63.8           \\ 
\midrule
Ours Model                                            & \textbf{66.8}  & \textbf{65.6}  \\ 
\bottomrule
\end{tabular}
\caption{Accuracy comparison between ours and other methods, where ``-'' means not available.}
\label{result}
\vspace{-.2in}
\end{table} 

\subsection{Results and Analysis}

\paragraph{Main Results.} We compared our model with several previous state-of-the-art methods and the methods using the same backbone as ours. The results are summarized in Table~\ref{result}. Observed that our model outperforms all the compared methods.\footnote{Note that we do not compare with the models utilizing stronger language models or external knowledge bases.} The BERT based model also shows stable superiority over the models based on other pre-trained language model such as ELMo and GPT. Despite the simplicity and few additional parameters of our model, it still outperforms other BERT based approaches. For instance, our model surpasses BERT Multi-task method by $1.8\%$ in accuracy, where BERT Multi-task is first trained on RACE and then fine-tuned on both OpenbookQA and ARC~\cite{clarkthink}.

\begin{wraptable}{r}{9cm}
\centering
\setlength\tabcolsep{1pt}
\small
\begin{tabular}{c c}
\toprule
Modification                                                    & Accuracy (\%) \\ 
\midrule
(1) w/o block fusion; w/o evidence filter                          & 60.0      \\ 
(2) w/o block fusion; evidence filter w/o constraints         & 63.8      \\ 
(3) w/o block fusion; evidence filter                             & 65.0      \\ 
(4) block fusion with same evidence filter                       & 64.0      \\ 
\midrule
block fusion with different evidence filter (ours)           & 65.6      \\
\bottomrule
\end{tabular}
\caption{The performance of our model and its ablations on OpenbookQA test set.}
\label{ablation}
\vspace{-.2in}
\end{wraptable}

\paragraph{Ablation Studies.} In Table~\ref{ablation}, we report the results of $5$ different ablation settings: (1) remove block fusion layer in equation \ref{block fusion} and evidence filter from our model; (2) remove block fusion layer and keep evidence filter without constraints; (3) remove block fusion layer and keep evidence filter with constraints; (4) keep both block fusion layer and evidence filter with constraints, all the blocks share the same evidence filter. According to the ablative results, we see that our full approach achieves the overall top performance. It demonstrates the effectiveness of evidence filter by comparing (1) with (2)-(4). In (2), we also test the trained model with three sets of shuffled options using different random seed and get varied results: 63.8\%, 63.6\% and 63\%, which exposes the problem mentioned in Section~\ref{sec:interaction_matrix}. However, all the experimental setups based on constraint-attached matrix are not affected. The results from (3), (4) and the last one (ours) suggest that block fusion sharing the same evidence filter performs worse than the model without block fusion, while performs better when different evidence filters are applied.

\paragraph{Analysis of Evidence Filter.} 
As discussed in Section~\ref{sec:interaction_matrix}, we expect the $\alpha$ and $\beta$ of the evidence filter should satisfy $\alpha\times\beta<0$. The values of $\alpha$ and $\beta$ for all blocks are $1.3432\pm0.0034$ and $-1.0680\pm0.0033$ respectively. \footnote{Full results are shown in supplementary material.} We can observe that the values of $\alpha$ and $\beta$ show opposite sign for all blocks. Another fact is that $\alpha$ are positive values and $\beta$ are negatives, which also represent the intra-interaction and inter-interaction. We can conclude that the parameters of the well-designed evidence filter are consistent with our intuition, which also provides an explanation for itself.

\section{Conclusions}\label{sec:conclusion}
We propose evidence filter to alleviate the effect of unrelated sentences and enhance the saliency of evidences potentially without human efforts. Results on OpenbookQA indicate the effectiveness of our method. Our future work is to enhance the evidence filter by more complex components.

\bibliographystyle{coling}
\bibliography{coling2020}

\newpage

\setcounter{section}{0}

\begin{center}
    \textbf{\Large Context Modeling with Evidence Filter for Multiple Choice Question Answering\\[.1cm]
    \textit{Supplementary Materials}}\\[.3cm]
    \large \textbf{Sicheng Yu}$^\clubsuit$,~~\textbf{Hao Zhang}$^{\vardiamondsuit,\varheartsuit}$,~~\textbf{Wei Jing}$^{\spadesuit,\varheartsuit}$,~~\textbf{Jing Jiang}$^\clubsuit$ \\
    $^\clubsuit$School of Information System, Singapore Management University, Singapore \\
    $^\vardiamondsuit$School of Computer Science and Engineering, Nanyang Technological University, Singapore \\
    $^\spadesuit$Institute of High Performance Computing, A*STAR, Singapore \\
    $^\varheartsuit$Institute for Infocomm Research, A*STAR, Singapore\\
    {\tt \{scyu.2018@phdcs.,jingjiang@\}smu.edu.sg\\
    \{26hzhang,21wjing\}@gmail.com}
\end{center}

\renewcommand\thesubsection{\Alph{subsection}}

\section{Derivation for Evidence Filter}
Firstly, we explain why evidence filter without constrains may cause different prediction when shuffling the options. For clear illustration, we use $\mathbf{R}\in{\mathbb{R}^{4\times4}}$ to represent a row (column) exchange matrix which has the following properties:
\begin{equation}
\begin{split}
    \sum_j{\mathbf{R}_{ij}}=1,\;\;\; \sum_i{\mathbf{R}_{ij}}=1,\;\;\;
    \mathbf{R}_{ij}=0\,\textrm{or}\,1.
\end{split}
\end{equation}
The output of block $k$ in BERT corresponding to a same sample with shuffled options is equivalent to the row exchange of $\text{BERT}_k[C;(Q;O_{i})]^4$, which can be written as $\mathbf{R}\cdot{\text{BERT}_k[C;(Q;O_{i})]^4}$. Shuffling the order of inputs to the evidence filter should be equivalent to first entering the evidence filter and then shuffling the output. The expected property of the evidence filter can be formulated as:
\begin{equation}
    \mathbf{A}\cdot{(\mathbf{R}\cdot{\text{BERT}_k[C;(Q;O_{i})]^4})}=\mathbf{R}\cdot{(\mathbf{A}\cdot{\text{BERT}_k[C;(Q;O_{i})]^4})}
\end{equation}
which is not satisfied if evidence filter is no-constrains attached.

Then we prove that evidence filter with constrains is about to completely solve this unexpected phenomenon.

\textbf{Proof:} We need to demonstrate that $\mathbf{A}\cdot{\mathbf{R}}=\mathbf{R}\cdot{\mathbf{A}}$. 
Recall the definition of $\mathbf{A}$ and $\mathbf{R}$.
The $i,j$ entry of $\mathbf{A}\cdot{\mathbf{R}}$ and $\mathbf{R}\cdot{\mathbf{A}}$ can be easily derived as following:
\begin{equation}
(\mathbf{A}\cdot{\mathbf{R}})_{ij}=\sum_k{\mathbf{A}_{ik}\cdot{\mathbf{R}_{kj}}=\alpha\mathbf{R}_{ij}}+\beta\sum_{k\neq{i}}\mathbf{R}_{kj}=(\alpha-\beta)\mathbf{R}_{ij}+\beta,
\end{equation}

\begin{equation}
(\mathbf{R}\cdot{\mathbf{A}})_{ij}=\sum_k{\mathbf{R}_{ik}\cdot{\mathbf{A}_{kj}}=\alpha\mathbf{R}_{ij}}+\beta\sum_{k\neq{j}}{\mathbf{R}_{ik}}=(\alpha-\beta)\mathbf{R}_{ij}+\beta,
\end{equation}
where $\alpha$ represents the diagonal values and $\beta$ denotes the rest parameters of evidence filter.

\section{Values of Parameters in Evidence Filter}
Full values of parameters, $\alpha$ and $\beta$, are shown in Table~\ref{value}.

\begin{table*}[ht]
\centering
\resizebox{\textwidth}{!}{
\begin{tabular}{c c c c c c c c c c c c c}
\toprule
Index & 1       & 2       & 3       & 4       & 5       & 6       & 7       & 8       & 9       & 10      & 11      & 12      \\ 
\midrule
$\alpha$          & 1.3418  & 1.3418  & 1.3418  & 1.3418  & 1.3418  & 1.3418  & 1.3408  & 1.3389  & 1.3447  & 1.3457  & 1.3447  & 1.3389  \\ 
$\beta$           & -1.0693 & -1.0693 & -1.0693 & -1.0693 & -1.0693 & -1.0693 & -1.0703 & -1.0723 & -1.0664 & -1.0664 & -1.0664 & -1.0723 \\ 
\midrule
Index & 13      & 14      & 15      & 16      & 17      & 18      & 19      & 20      & 21      & 22      & 23      & 24      \\ 
\midrule
$\alpha$          & 1.3379  & 1.3457  & 1.3369  & 1.3477  & 1.3467  & 1.3457  & 1.3398  & 1.3467  & 1.3477  & 1.3477  & 1.3486  & 1.3408  \\ 
$\beta$           & -1.0723 & -1.0654 & -1.0742 & -1.0635 & -1.0645 & -1.0654 & -1.0713 & -1.0654 & -1.0635 & -1.0635 & -1.0625 & -1.0713 \\ 
\bottomrule
\end{tabular}}
\caption{Values of $\alpha$ and $\beta$ in evidence filter for each block of model after training.}
\label{value}
\end{table*}

\end{document}